\documentclass[letterpaper]{article} 
\usepackage{aaai2026}  
\usepackage{times}  
\usepackage{helvet}  
\usepackage{courier}  
\usepackage[hyphens]{url}  
\usepackage{graphicx} 
\usepackage{natbib}  
\usepackage{caption} 
\usepackage{algorithm}
\usepackage{algorithmic}

\usepackage{amsfonts}       
\usepackage{amssymb} 
\usepackage[table]{xcolor}
\usepackage{booktabs}  
\usepackage{amsmath}  
\usepackage{multirow}
\usepackage{newfloat}
\usepackage{listings}

\DeclareCaptionStyle{ruled}{labelfont=normalfont,labelsep=colon,strut=off} 
\floatstyle{ruled}
\newfloat{listing}{tb}{lst}{}
\floatname{listing}{Listing}
\title{Shared \& Domain Self-Adaptive Experts with Frequency-Aware Discrimination for Continual Test-Time Adaptation}
\author {
    Jianchao Zhao\textsuperscript{\rm 1},
    Chenhao Ding\textsuperscript{\rm 3},
    SongLin Dong\textsuperscript{\rm 2}\thanks{Corresponding author.},
    Jiangyang Li\textsuperscript{\rm 1},\\
    Qiang Wang\textsuperscript{\rm 1},
    Yuhang He\textsuperscript{\rm 1},
    Yihong Gong\textsuperscript{\rm 1,2}
}
\affiliations {
    \textsuperscript{\rm 1} National Key Laboratory of Human-Machine Hybrid Augmented Intelligence,\\National Engineering Research Center for Visual Information and Applications,\\Institute of Artificial Intelligence and Robotics, Xi'an Jiaotong University \\
    \textsuperscript{\rm 2} Faculty of Computility Microelectronics, Shenzhen University of Advanced Technology\\
    \textsuperscript{\rm 3} School of Software Engineering, Xi'an Jiaotong University
    \\
    \{zhaojianchao, dch225739, ljy1021,qwang\}@stu.xjtu.edu.cn, dongsl@suat-sz.edu.cn,\\
    heyuhang@xjtu.edu.cn, ygong@mail.xjtu.edu.cn
}

\begin{document}

\maketitle

\begin{abstract}
This paper focuses on the Continual Test-Time Adaptation (CTTA) task, 
aiming to enable an agent to continuously adapt to evolving target domains while retaining previously acquired domain knowledge for effective \textbf{reuse} when those domains \textbf{reappear}.
Existing shared-parameter paradigms struggle to balance adaptation and forgetting, leading to decreased efficiency and stability. To address this, we propose a frequency-aware shared and self-adaptive expert framework, consisting of two key components: (i) a dual-branch expert architecture that extracts general features and dynamically models domain-specific representations, effectively reducing cross-domain interference and repetitive learning cost; and (ii) an online Frequency-aware Domain Discriminator (FDD), which leverages the robustness of low-frequency image signals for online domain shift detection, guiding dynamic allocation of expert resources for more stable and realistic adaptation. Additionally, we introduce a Continual Repeated Shifts (CRS) benchmark to simulate periodic domain changes for more realistic evaluation. 
Experimental results show that our method consistently outperforms existing approaches on both classification and segmentation CTTA tasks under standard and CRS settings, with ablations and visualizations confirming its effectiveness and robustness. Our code is available at https://github.com/ZJC25127/Domain-Self-Adaptive-CTTA.git
\end{abstract}


\section{Introduction}
Test-Time Adaptation (TTA)~\cite{liang2025comprehensive,han2025unleashing} enables pretrained models to adapt to unknown data distributions during inference, demonstrating significant potential for practical applications~\cite{dong2024brain,gong2024preface,peng2024global,peng2025cia,wang2024non}. Traditional TTA methods primarily focus on static target domains that assume a fixed distribution. However, in real-world scenarios such as autonomous driving, models encounter environmental variations caused by changes in weather, sensors, or location over time, requiring agents to adapt to these evolving conditions continuously. Motivated by this, Continual Test-Time Adaptation (CTTA)~\cite{wang2022continual}, which aligns closely with practical demands, has attracted increasing attention by enabling models to effectively handle a sequence of continually changing unknown distributions.

\begin{figure}[t]           
  \centering
  \includegraphics[width=\columnwidth]{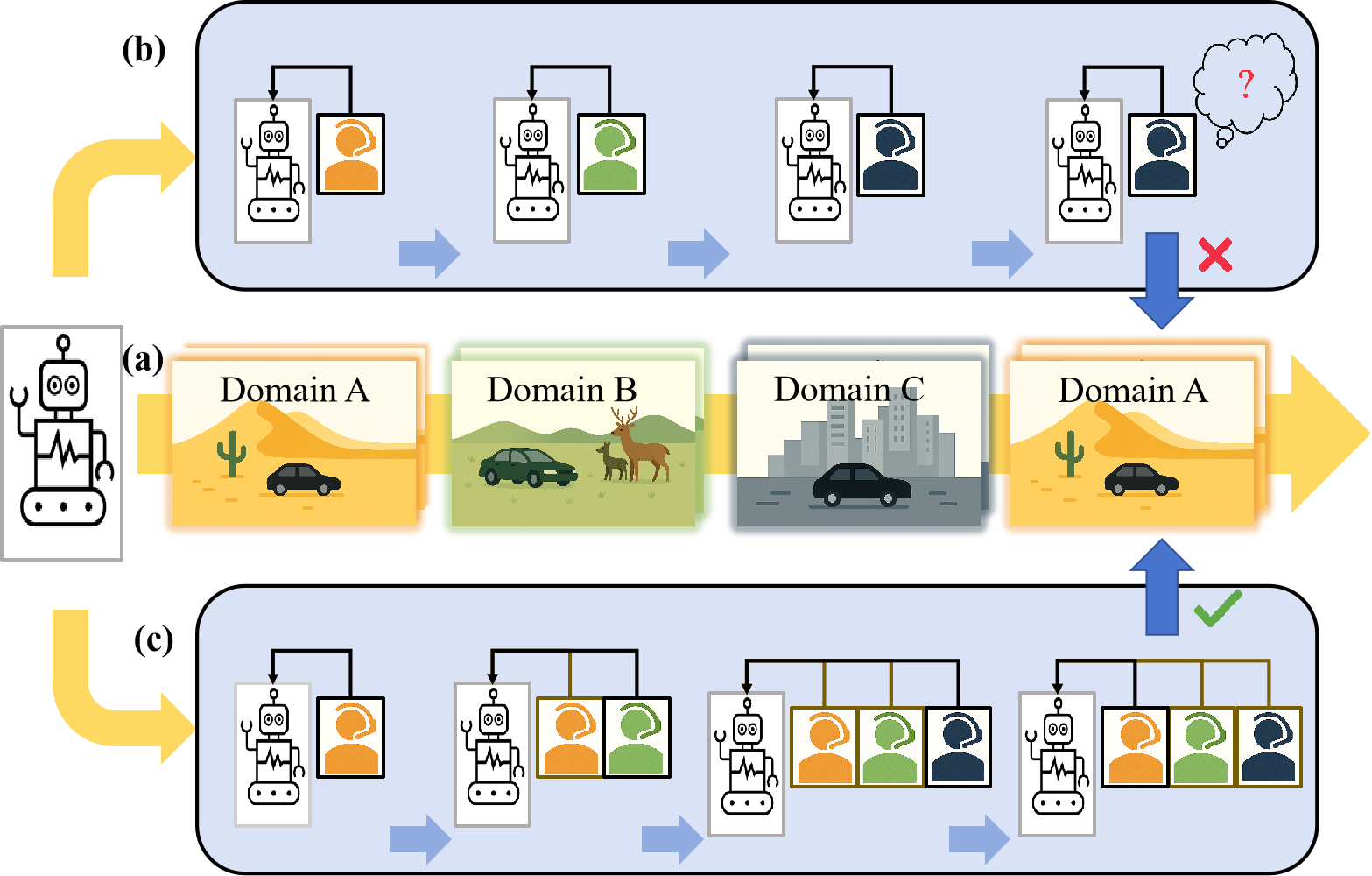}
  \caption{\textbf{A New Benchmark and Comparison of Methods for CTTA.}
  (a) Our proposed CRS (Continual Repeated Shifts) benchmark simulates realistic, cyclic domain shifts at test time.
  (b) Existing shared-parameter methods struggle to balance adaptation and memory, leading to forgetting and redundant relearning when domains recur.
  (c) Our method adds domain self-adaptive experts, enabling knowledge retention and fast adaptation to recurring domains.
}
  \label{fig:teaser}
\end{figure}

Existing mainstream CTTA methods are divided into two categories: (i) model-based paradigm: this paradigm~\cite{dobler2023robust,liu2024continual} updates the entire or partial modules of the model during testing by generating pseudo-labels and leveraging consistency or self-supervised objectives to achieve online adaptation to new domains; (ii) parameter-tuning based: this paradigm~\cite{vida,gan2023decorate,song2023ecotta} is based on parameter-efficient fine-tuning (PEFT) strategies~\cite{hu2021lora,ding2025sulora}, which introduce learnable prompts or LoRA modules into the model to adapt to continuously changing target domain knowledge. Methods of paradigm (i) usually incur high computational costs and lack fine-grained modeling of domain discrepancies, leading to low adaptation efficiency or unstable performance prone to forgetting during continual adaptation. Therefore, we focus on an in-depth study of paradigm (ii).

Paradigm (ii) methods typically achieve continual adaptation to unseen domains by relying on shared tuning parameters, which have inherent limitations. Consider an autonomous-driving vision model~(Fig.~\ref{fig:teaser}) which must continually recognize and interpret road scenes: over time it traverses recurring environments—deserts, grasslands, urban areas. Ideally, the model should adapt rapidly to new environments while retaining domain-specific knowledge to reuse when the same environment reappears. However, these methods often suffer from forgetting prior domain knowledge while adapting to new domains, leading to an inherent trade-off between adaptation and memory retention~\cite{dong2024ceat,dong2025analogical,wang2025dualcp}. When prior domains reappear, the model is forced to relearn what it has already forgotten, hurting efficiency and stability. Although some approaches~\cite{vida,gan2023decorate} attempt to disentangle domain-shared and domain-specific features within the shared parameter space, the training process still relies on a unified set of parameters. As a result, the underlying conflict between adaptation and memory retention remains unresolved. 

To address this, we propose a frequency-discriminative shared \& self-adaptive domain expert model. It has two key components: (i) a dual-branch expert architecture; and (ii) an online frequency-aware domain discriminator (FDD). Specifically, the first module comprises a shared expert and domain self-adaptive experts. The shared expert continuously adapts to extract task-relevant general features, ensuring stable transfer of discriminative knowledge across domains, while domain self-adaptive experts are dynamically introduced upon detecting new domains to model domain-specific features, mitigating cross-domain interference and reducing redundant learning. To achieve accurate domain shift detection and robust expansion, the FDD module leverages low-frequency image signals for real-time domain drift perception. It adaptively allocates the most suitable expert resources for new domains while preserving existing domain knowledge, enabling more realistic and stable continuous test-time adaptation.

Furthermore, current CTTA experiments learn each domain only once, ignoring realistic repeated shifts (e.g., rain$\rightarrow$sun$\rightarrow$rain). We therefore propose a more realistic benchmark, Continual Repeated Shifts (CRS), constructed by cyclically combining four ImageNet-derived datasets to simulate periodic domain repetitions (see Fig.\ref{fig:teaser} (b)), with extended versions ImageNet+ and ImageNet++. Compared to conventional CTTA tasks, CRS shows larger domain discrepancies and periodic distribution shifts, more faithfully replicating dynamic real-world environments. 

The key contributions are as follows: we identify the limitations of existing paradigms in managing the trade-off between forgetting and adaptation, and propose a domain shared \& self-adaptive expert with frequency-aware discrimination to address this challenge. We design a realistic CTTA benchmark, Continual Repeated Shifts (CRS), which captures the recurrent nature of domain shifts in real-world scenarios. Extensive experiments on classification (general and CRS settings) and semantic segmentation demonstrate that our method consistently outperforms others.

\section{Related Work}
\paragraph{Parameter-Efficient Fine-Tuning.} Parameter-efficient fine-tuning (PEFT) methods adapt large pre-trained models to downstream tasks with few trainable parameters and low compute. Representative approaches include Adapters~\cite{houlsby2019parameter}, Low‑Rank Adaptation (LoRA)~\cite{hu2021lora}, and Prompt Tuning~\cite{jia2022visual}. Owing to its plug‑and‑play insertion and tiny memory footprint, LoRA is widely adopted in practice. Building on LoRA, Mixture‑of‑Experts (MoE)~\cite{shazeer2017outrageously,wang2022adamix} splits the low‑rank update into multiple experts and employs a gating mechanism to activate only a few per input, preserving scalability while reducing inference FLOPs. Recent work includes DA‑MoE~\cite{aghda2024damo}, which allocates a variable number of experts to each token based on salience for fine‑grained compute–confidence trade‑offs, and DeepSeekMoE~\cite{zhang2024deepseekmoe}, which introduces fine‑grained expert partitioning with shared sub‑experts to cut redundancy and boost specialization. Inspired by DeepSeekMoE, we propose a hybrid architecture combining shared and self-adaptive experts, well-suited to tasks with complex, continuous domain drift.

\paragraph{Continual Test-Time Adaptation.} Continual Test-Time Adaptation (CTTA) extends the TTA paradigm by addressing distribution shifts across a sequence of unseen target domains. CoTTA~\cite{wang2022continual} first addresses this task by introducing a teacher‑averaged pseudo‑labeling strategy and stochastic parameter restoration. Subsequent methods build on this foundation. RMT~\cite{dobler2023robust} replaces standard cross-entropy with symmetric cross-entropy to improve gradient stability and combines it with contrastive learning to preserve feature alignment with the source. EcoTTA~\cite{song2023ecotta} addresses memory efficiency by introducing meta-networks that adapt lightweight layers and employ self-distillation to preserve source knowledge. BECoTTA~\cite{lee2024becotta} proposes a mixture-of-experts framework that leverages domain-adaptive routing to minimize parameter updates while maintaining adaptation performance. Meanwhile, VDP~\cite{gan2023decorate} and VIDA~\cite{vida} attempt to decouple task-relevant and domain-specific knowledge through prompt-based or adapter-based mechanisms. Despite these advances, existing methods still struggle to strike an optimal balance between rapid adaptation and forgetting mitigation due to shared parameters, and also struggle to cope with more realistic and complex scenarios.

\begin{figure*}[t]      
  \centering
  \includegraphics[width=\textwidth]{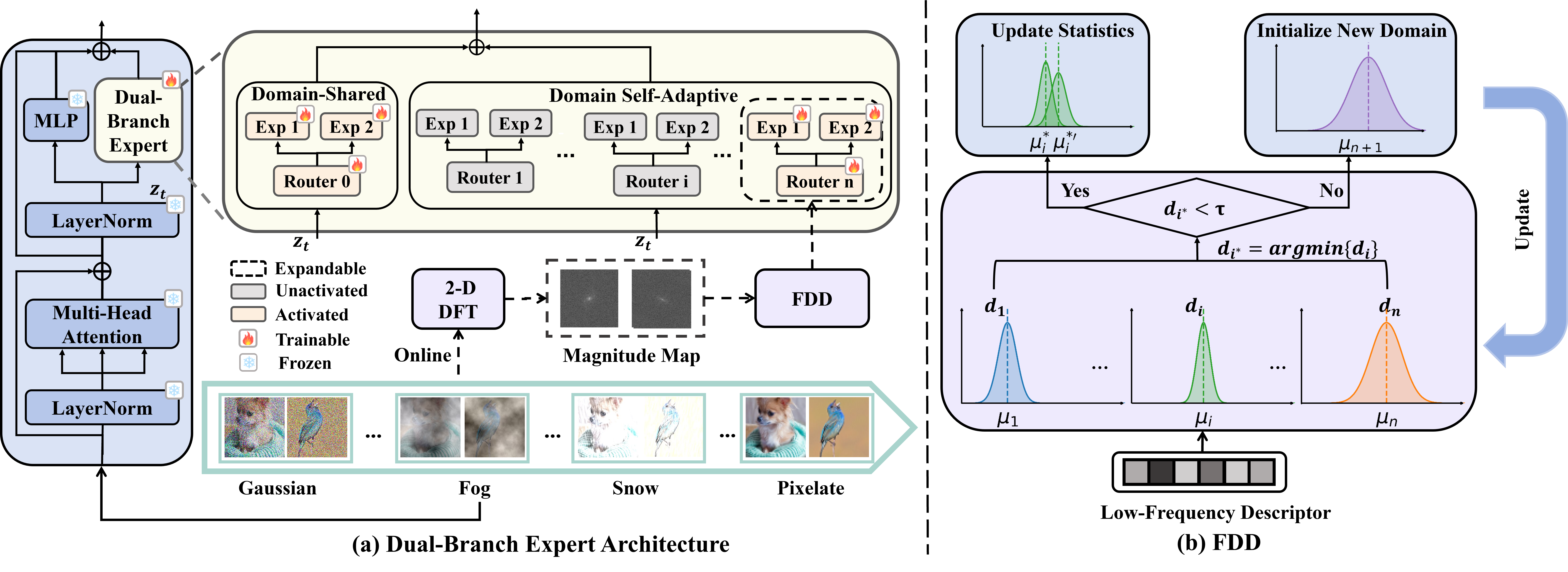}
  \caption{Overall framework.
(a) The Dual-Branch Expert Architecture combines a shared expert with a dynamically expandable set of domain self-adaptive experts. The appropriate domain self-adaptive expert is selected based on input, guided by the online Frequency-aware Domain Discriminator (FDD).
(b) FDD detects domain shifts using low-frequency features from 2D Fourier-transformed inputs, updating domain statistics or initializing new experts when unseen domains are encountered.}
  \label{fig:expamoe_overview}
\end{figure*}

\section{Method}
\subsection{Preliminary}
\label{sec:Preliminary}
\paragraph{Problem Formulation.}
In Continual Test-Time Adaptation (CTTA), the model $q_\theta(y|x)$ is first pre-trained on a labeled source domain $\mathcal{D}_S = \{(x_s, y_s)\}$. After deployment, the model is adapted to a sequence of target domains $\{\mathcal{D}_{T_i}\}_{i=1}^{n}$, where $n$ denotes the number of continual shifts, and the distributions of the target domains $\mathcal{D}_{T_1}, \mathcal{D}_{T_2}, \ldots, \mathcal{D}_{T_n}$ evolve over time. The CTTA protocol follows three key assumptions: (1) access to the source domain $\mathcal{D}_S$ is strictly prohibited after deployment, (2) each target domain sample $x \in \mathcal{D}_T$ can be observed only once during adaptation, and (3) no ground-truth labels are available for the target domains. Under these constraints, the objective of CTTA is to adapt the pre-trained model $q_\theta$ to evolving target domains while maintaining performance and preserving recognition ability on previously seen distributions.


\paragraph{Mixture-of-Experts.}
A MoE module consists of a router and $M$ experts ${E_1, \ldots, E_M}$. Given an input $\mathbf{z} \in \mathbb{R}^D$ (where $D$ is the feature dimension), the router computes gating scores $\mathbf{g} = W_r \mathbf{z} + \mathbf{b}_r$ using learnable parameters $W_r \in \mathbb{R}^{M \times D}$ and $\mathbf{b}_r \in \mathbb{R}^M$, and normalizes them via softmax to obtain mixture weights $\boldsymbol{\alpha} \in \mathbb{R}^M$.

Each expert $E_i$ is a two-layer low-rank network:
\begin{equation}
E_i(\mathbf{z}) = \sigma(\mathbf{z} W^{\text{down}}_i) W^{\text{up}}_i,
\end{equation}
where $W^{\text{down}}_i \in \mathbb{R}^{D \times r}$ and $W^{\text{up}}_i \in \mathbb{R}^{r \times D}$ are learnable projection matrices, and $r \ll D$ is the bottleneck dimension. The activation function $\sigma(\cdot)$ is typically ReLU or GELU.

The final MoE output is a weighted sum of expert outputs:
\begin{equation}
\mathrm{MoE}(\mathbf{z}) = \sum_{i=1}^{M} \alpha_i \cdot E_i(\mathbf{z}).
\end{equation}




\subsection{Dual-Branch Expert Architecture}
\label{sec:DBE}
As illustrated in Fig.~\ref{fig:expamoe_overview}, our dual‑branch architecture comprises (i) a \textit{domain-shared expert branch} that encodes domain‑invariant task knowledge and (ii) a \textit{domain self-adaptive expert branch} that grows on demand to capture domain‑specific shifts.

\paragraph{Domain-Shared Expert Branch.}  
Given the input token sequence \(\mathbf{Z} \in \mathbb{R}^{B \times N \times D}\), where \(B\) is the batch size, \(N\) is the number of tokens per sample. we feed it into a fixed Mixture-of-Experts (MoE) layer shared across all domains:
\begin{equation}
\mathbf{Y}_{\text{shared}} = \mathrm{MoE}_{\text{shared}}(\mathbf{Z}).
\label{eq:shared_output}
\end{equation}
This branch facilitates knowledge transfer and supports generalization by preventing the overwriting of task-critical representations during continual adaptation.

\paragraph{Domain Self-Adaptive Expert Branch.}  

To handle domain-specific variations, we introduce an expandable pool of MoE modules in domain self-adaptive expert branch. Input token sequence \(\mathbf{Z}\) is processed by one of the domain self-adaptive expert modules: \( \{\mathrm{MoE}^{(1)}_{\text{domain}}, \mathrm{MoE}^{(2)}_{\text{domain}}, \ldots, \mathrm{MoE}^{(n)}_{\text{domain}}\} \).

Each expert module \(\mathrm{MoE}^{(i)}_{\text{domain}}\) is responsible for modeling a previously observed domain and consists of a router \(G^{(i)}\) and a set of lightweight experts \(\{E_1^{(i)}, \ldots, E_M^{(i)}\}\). 
The expert module is dynamically selected based on the current input using the online Frequency-Aware Domain Discriminator (FDD, see Section~\ref{sec:fodd}):
\begin{equation}
\mathbf{Y}_{\text{domain}} = \mathrm{MoE}^{(i^*)}_{\text{domain}}(\mathbf{Z}),
\label{eq:domain_output_new}
\end{equation}
where \(i^*\) denotes the selected expert module. If the current input distribution does not match any existing module, a new module \(\mathrm{MoE}^{(n+1)}_{\text{domain}}\) is instantiated and added to the pool.

\paragraph{Fusion of Dual Outputs.}  
To integrate both generalizable and domain-adaptive knowledge, we combine the outputs of the two branches using a residual fusion mechanism:
\begin{equation}
\mathbf{Z}_{\text{out}} = \mathbf{Z} + \lambda \cdot \mathbf{Y}_{\text{shared}} + (1 - \lambda) \cdot \mathbf{Y}_{\text{domain}},
\label{eq:final_output_simple}
\end{equation}
where \(\lambda \in [0, 1]\) is a hyperparameter controls the balance between domain-shared and domain-specific representations.

\subsection{Online Frequency-Aware Domain Discriminator}
\label{sec:fodd}
Efficient domain‑shift detection is critical to our CTTA framework; therefore we introduce online Frequency-Aware Domain Discriminator (FDD), a lightweight and training‑free module that leverages frequency statistics to detect shifts and route inputs to the appropriate domain self-adaptive expert.

\paragraph{Bayesian Domain Posterior Estimation and Decision Rule.}
Given an input image $x$, we obtain its low‑frequency descriptor
$z = f(x) \in \mathbb{R}^{d_f}$, where $f(\cdot)$ is the low‑frequency feature extractor detailed at the end of this subsection, and \(d_f\) denotes the dimension of the resulting frequency descriptor. Suppose the model has so far identified $K$ distinct domains during adaptation, let $y \in \{1, \dots, K\}$ denote the latent domain label. Inspired by classical Gaussian discriminant analysis \citep{hastie1996discriminant}, we assume that the low-frequency embeddings corresponding to each domain follow a multivariate Gaussian distribution:
\begin{equation}
p(z \mid y=i) = \mathcal{N}(z \mid \mu_i, \Sigma_i),
\end{equation}
where $\mu_i \in \mathbb{R}^{d_f}$ and $\Sigma_i \in \mathbb{R}^{{d_f} \times {d_f}}$ are the estimated mean and covariance of domain $i$, respectively. We further assume a uniform prior over domains, i.e., $P(y=i) = 1/K$. Then, the posterior probability that a sample belongs to domain $i$ is given by Bayes' rule:
\begin{equation} 
\begin{aligned}
P(y=i \mid z)
  &= \frac{p(z \mid y=i)}
          {\sum_{j=1}^{K} p(z \mid y=j)} \\[4pt]  
  &= \frac{\exp\!\bigl(-\tfrac12 m_i(z)\bigr)/\sqrt{\det\Sigma_i}}
          {\displaystyle\sum_{j=1}^{K}
            \exp\!\bigl(-\tfrac12 m_j(z)\bigr)/\sqrt{\det\Sigma_j}},
\end{aligned}
\end{equation}
where $m_i(z)$ denotes the Mahalanobis distance under a shrinkage-regularized covariance, and \(\det \Sigma\) is the determinant of the covariance matrix. More detail can be found in Appendix~\ref{app:mahalanobis}.

To assign a domain label to an incoming batch of $B$ samples $\{x_b\}_{b=1}^B$, we compute the average embedding:
\begin{equation}
\bar{z}^{(0)}_t = \frac{1}{B} \sum_{b=1}^{B} f(x_b),
\end{equation}
and select the domain with the highest posterior probability, which is equivalent (see Appendix~\ref{app:posterior-mahalanobis}) to choosing the domain with the smallest Mahalanobis distance:
\begin{equation}
i^* = \arg\min_i m_i\big(\bar{z}^{(0)}_t\big).
\end{equation}
If the minimum distance $\min_i m_i(\bar{z}^{(0)}_t)$ exceeds a pre-defined threshold $\tau$, we infer that the current batch likely originates from a previously unseen domain, and a new domain self-adaptive expert module is initialized. Otherwise, the batch is assigned to the expert module corresponding to the closest known domain $i^*$.

\begin{table*}[t]
\centering
\small              
\setlength\tabcolsep{1pt}  
\begin{tabular}{c|c|ccccccccccccccc|cc}
\toprule
Method & REF &
\rotatebox[origin=c]{50}{Gaussian} & \rotatebox[origin=c]{50}{Shot} &
\rotatebox[origin=c]{50}{Impulse} & \rotatebox[origin=c]{50}{Defocus} &
\rotatebox[origin=c]{50}{Glass} & \rotatebox[origin=c]{50}{Motion} &
\rotatebox[origin=c]{50}{Zoom} & \rotatebox[origin=c]{50}{Snow} &
\rotatebox[origin=c]{50}{Frost} & \rotatebox[origin=c]{50}{Fog} &
\rotatebox[origin=c]{50}{Brightness} & \rotatebox[origin=c]{50}{Contrast} &
\rotatebox[origin=c]{50}{Elastic} & \rotatebox[origin=c]{50}{Pixelate} &
\rotatebox[origin=c]{50}{JPEG} & Mean$\downarrow$ & Gain \\
\midrule
Source & ICLR2021  & 53.0 & 51.8 & 52.1 & 68.5 & 78.8 & 58.5 & 63.3 & 49.9 & 54.2 & 57.7 & 26.4 & 91.4 & 57.5 & 38.0 & 36.2 & 55.8 & 0.0 \\
Pseudo-label & ICML2013 & 45.2 & 40.4 & 41.6 & 51.3 & 53.9 & 45.6 & 47.7 & 40.4 & 45.7 & 93.8 & 98.5 & 99.9 & 99.9 & 98.9 & 99.6 & 61.2 & -5.4 \\
TENT-continual & ICLR2021 & 52.2 & 48.9 & 49.2 & 65.8 & 73.0 & 54.5 & 58.4 & 44.0 & 47.7 & 50.3 & 23.9 & 72.8 & 55.7 & 34.4 & 33.9 & 51.0 & +4.8 \\
CoTTA & CVPR2022 & 52.9 & 51.6 & 51.4 & 68.3 & 78.1 & 57.1 & 62.0 & 48.2 & 52.7 & 55.3 & 25.9 & 90.0 & 56.4 & 36.4 & 35.2 & 54.8 & +1.0 \\
VDP & AAAI2023 & 52.7 & 51.6 & 50.1 & 58.1 & 70.2 & 56.1 & 58.1 & 42.1 & 46.1 & 45.8 & 23.6 & 70.4 & 54.9 & 34.5 & 36.1 & 50.0 & +5.8 \\
ViDA & ICLR2024 & 47.7 & 42.5 & 42.9 & 52.2 & 56.9 & 45.5 & 48.9 & 38.9 & 42.7 & 40.7 & 24.3 & 52.8 & 49.1 & 33.5 & 33.1 & 43.4 & +12.4 \\
ADMA & CVPR2024 & \textbf{46.3} & \textbf{41.9} & 42.5 & 51.4 & 54.9 & 43.3 & \textbf{40.7} & \textbf{34.2} & \textbf{35.8} & 64.3 & 23.4 & 60.3 & \textbf{37.5} & 29.2 & 31.4 & 42.5 & +13.3 \\
\rowcolor{gray!20}
\textbf{Ours} & \textbf{Proposed} & 47.7 & 45.1 & \textbf{42.2} & \textbf{46.6} & \textbf{49.7} & \textbf{42.8} & 46.5 & 35.0 & 38.0 & \textbf{35.2} & \textbf{21.6} & \textbf{51.7} & 43.5 & \textbf{26.9} & \textbf{31.0} & \textbf{40.2} & \textbf{+15.6} \\
\bottomrule
\end{tabular}
\caption{Classification error rates (\%) for the ImageNet-to-ImageNet-C CTTA task. \textit{Mean} is the average error across 15 corruption types. \textit{Gain} shows accuracy improvement over the source model.}
\label{tab:ImageNet-C}
\end{table*}

\paragraph{Robust Online Update of Domain Statistics.}
In CTTA, samples arrive sequentially, making it impossible to observe the full distribution of any domain at once. To address this, we adopt an incremental estimation strategy that is both memory-efficient and noise-robust, enabling online updates of domain statistics based on the current batch.

To this end, we adopt a soft-assignment strategy, assigning each sample $x_b$ a normalized responsibility weight computed from its Gaussian probability density under the selected domain’s distribution:
\begin{equation}
w_b = \frac{\exp\left[-\frac{1}{2} m_{i^*}(f(x_b))\right]}{\sum_{j=1}^{B} \exp\left[-\frac{1}{2} m_{i^*}(f(x_j))\right]}, \qquad \sum_{b=1}^{B} w_b = 1.
\label{eq:soft_weights}
\end{equation}
These normalized weights effectively down-weight outlier samples with poor domain fit, acting as a robust mechanism to stabilize parameter updates.

Assuming the current cumulative weight (or effective batch count) for domain $i^*$ is $c_{i^*}$, and its corresponding sample statistics are summarized by the mean $\mu_{i^*}$ and covariance $\Sigma_{i^*}$. we define the weighted complete-data log-likelihood for the batch as:
\begin{equation}
\mathcal{L}(\mu, \Sigma) = c_{i^*} \log \mathcal{N}(\mu_{i^*} \mid \mu, \Sigma) + \sum_{b=1}^{B} w_b \log \mathcal{N}(z_b \mid \mu, \Sigma),
\label{eq:likelihood}
\end{equation}
where \( z_b = f(x_b) \) is the low-frequency
feature of sample \( x_b \).
Maximizing $\mathcal{L}$ with respect to $\mu$ and $\Sigma$ yields the following closed-form updates (see Appendix~\ref{app:mle-update}) for detailed derivation):
\begin{align}
\mu_{i^*}^{\text{new}} &= \frac{c_{i^*} \mu_{i^*} + \sum_{b=1}^{B} w_b f(x_b)}{c_{i^*} + 1}, \label{eq:mu_update}\\
\Sigma_{i^*}^{\text{new}} &= \frac{c_{i^*} \Sigma_{i^*} + \sum_{b=1}^{B} w_b (f(x_b) - \mu_{i^*})(f(x_b) - \mu_{i^*})^\top}{c_{i^*} + 1}.
\label{eq:sigma_update}
\end{align}
Finally, the domain's sufficient statistics are updated as:
\begin{equation}
c_{i^*} \leftarrow c_{i^*} + 1, \qquad \mu_{i^*} \leftarrow \mu_{i^*}^{\text{new}}, \qquad \Sigma_{i^*} \leftarrow \Sigma_{i^*}^{\text{new}}.
\label{eq:final_update}
\end{equation}

When a new domain is detected (i.e., no existing domain achieves sufficiently high posterior confidence), we initialize the new domain’s statistics from the current batch. Specifically, we use the average embedding $\bar{z}^{(0)}_t$ as the initial mean, and a diagonal covariance scaled by a constant variance $\sigma_0^2$:
\begin{equation}
c_{\text{new}} = 1, \qquad \mu_{\text{new}} = \bar{z}^{(0)}_t, \qquad \Sigma_{\text{new}} = \sigma_0^2 I.
\end{equation}

This corresponds to the maximum a posteriori estimate under a Gaussian–Normal-Inverse-Wishart prior with a single observation, ensuring a well-posed initialization for the new domain.

\paragraph{Low-Frequency Feature Extractor.} 
Prior studies~\cite{yin2019fourier,wang2020towards,xie2022masked} indicate that low-frequency image features, encoding global statistics, are highly sensitive to domain shifts. Thus, we adopt low-frequency spectral features as lightweight indicators for domain discrimination.

Given an RGB image $x\in\mathbb{R}^{H\times W\times 3}$, we first convert it to a single–channel grayscale image
$x_{\text{gray}}\in\mathbb{R}^{H\times W}$. The two–dimensional discrete Fourier transform (2-D DFT) of $x_{\text{gray}}$ is then computed as:
\begin{equation}
F(u,v)=
\sum_{m=0}^{H-1}\sum_{n=0}^{W-1}
x_{\text{gray}}(m,n)\;
e^{-\,j\,2\pi\bigl(\tfrac{u\,m}{H}+\tfrac{v\,n}{W}\bigr)},
\label{eq:dft}
\end{equation}
where $F(u,v)\in\mathbb{C}^{H\times W}$ is the complex-valued spectrum, $j=\sqrt{-1}$ is the imaginary unit, and $0\leq u<H$, $0\leq v<W$.

Once the complex spectrum \( F(u,v) \) is obtained, we shift the DC component to the center of the frequency plane using a standard frequency-shift operation. The magnitude of the resulting complex spectrum is then taken to obtain a real-valued, centered magnitude map \( M \in \mathbb{R}^{H \times W} \).
Let the vertical and horizontal coordinates of the map center be $c_r=\lfloor H/2\rfloor$ and $c_c=\lfloor W/2\rfloor$, we keep only the low-frequency part by cropping a square patch of side length $L=2l+1$ around that centre:
\begin{equation}
f_{\text{low}} = M[c_r\!-\!l : c_r\!+\!l,\; c_c\!-\!l : c_c\!+\!l],
\label{eq:crop}
\end{equation}
where $l$ is a predefined frequency radius that controls the size of the low-frequency region. Finally, the cropped low-frequency patch $f_{\text{low}}$ is flattened into a vector $z \in \mathbb{R}^{L^2}$, which serves as the compact low-frequency descriptor.

\subsection{Optimization Objective}
\label{sec:optimization}
Following prior work in test-time adaptation, we minimize the entropy of model predictions for target samples, and update our expert modules only when the prediction entropy is below a threshold $\kappa$:
\begin{equation}
\mathcal{L}_{\text{TTA}} = 
- \mathbf{1}\left\{ \mathcal{H}(\hat{y}) < \kappa \right\}
\sum_{k=1}^{C} \hat{y}_k \log \hat{y}_k,
\end{equation}
where $\hat{y} = q_\theta(x)$ is the model’s predicted probability vector for input $x$, $C$ is the number of classes, and $\mathbf{1}\{\cdot\}$ denotes the indicator function.

\begin{table*}             

\centering
\setlength{\tabcolsep}{2.5pt}      
{\footnotesize 
\begin{tabular}{c|c|ccccc|ccccc|ccccc|c|c|c}
\toprule
\multicolumn{2}{c|}{Time} & \multicolumn{15}{c}{$t$ \makebox[10cm]{\rightarrowfill}} \\
\midrule
\multicolumn{2}{c|}{Round} &
\multicolumn{5}{c|}{1} & \multicolumn{5}{c|}{2} & \multicolumn{5}{c|}{3} &
\multirow{2}{*}{Mean$\downarrow$} & \multirow{2}{*}{RF$\downarrow$} & \multirow{2}{*}{Gain} \\
\cmidrule{1-17}
Method & REF & V2 & A & S & R & Mean$\downarrow$ &
V2 & A & S & R & Mean$\downarrow$ &
V2 & A & S & R & Mean$\downarrow$ & & \\
\midrule
Source & ICLR2021
& 44.8 & 61.3 & 58.4 & 43.0 & 51.2
& 44.8 & 61.3 & 58.4 & 43.0 & 51.2
& 44.8 & 61.3 & 58.4 & 43.0 & 51.2
& 51.2 & 0.0 & / \\

CoTTA & CVPR2022
& 44.7 & 61.3 & 58.0 & 43.0 & 51.0
& 44.6 & 59.7 & 57.5 & 42.3 & 50.5
& 44.5 & 58.6 & 56.9 & 42.1 & 50.1
& 50.2 &-0.9 & +1.0 \\

ViDA & ICLR2024
& 44.0 & \textbf{51.5} & \textbf{51.5} & 41.2 & 47.0
& 44.6 & 51.4 & 63.5 & 44.8 & 53.2
& 45.2 & 52.1 & 67.9 & 46.8 & 55.8
& 52.0 & +5.0 & -0.8 \\

\rowcolor{gray!20}
\textbf{Ours} & \textbf{Proposed}
& \textbf{43.9} & 52.2 & 52.0 & \textbf{38.4} & \textbf{46.5}
& \textbf{44.2} & \textbf{50.4} & \textbf{49.2} & \textbf{33.2} & \textbf{44.0}
& \textbf{44.1} & \textbf{49.6} & \textbf{47.9} & \textbf{32.1} & \textbf{43.1}
& \textbf{44.5} & \textbf{-3.4} & \textbf{+6.7} \\
\bottomrule
\end{tabular}
}
\caption{Classification error rates (\%) for the ImageNet-to-ImageNet+ CTTA task.}
\label{tab:ImageNet+}
\end{table*}

\begin{table*}[tb]                 
\centering

\setlength{\tabcolsep}{2.5pt}        
{\footnotesize 
\begin{tabular}{c|c|ccccc|ccccc|ccccc|c|c|c}
\toprule
\multicolumn{2}{c|}{Time} & \multicolumn{15}{c}{$t$ \makebox[10cm]{\rightarrowfill}} \\
\midrule
\multicolumn{2}{c|}{Round} &
\multicolumn{5}{c|}{1} & \multicolumn{5}{c|}{2} & \multicolumn{5}{c|}{3} &
\multirow{2}{*}{Mean$\downarrow$} & \multirow{2}{*}{RF$\downarrow$} & \multirow{2}{*}{Gain} \\
\cmidrule{1-17}
Method & REF & V2 & A & S & R & Mean$\downarrow$ &
V2' & A' & S' & R' & Mean$\downarrow$ &
V2'' & A'' & S'' & R'' & Mean$\downarrow$ & & \\
\midrule
Source & ICLR2021
& 27.1 & 61.3 & 58.3 & 43.2 & 48.4
& 15.8 & 61.3 & 58.3 & 42.6 & 45.8
& 19.7 & 61.3 & 58.6 & 43.3 & 46.9
& 47.0 & -1.5  & / \\

CoTTA & CVPR2022
& 27.1 & 61.3 & 58.1 & 43.1 & 48.3
& 15.6 & 60.7 & 57.9 & 42.8 & 45.5
& 19.5 & 59.8 & 57.3 & 41.7 & 45.8
& 46.5 & -2.5  & +0.5 \\

ViDA & ICLR2024
& 26.0 & \textbf{51.4} & \textbf{52.0} & 40.5 & \textbf{43.4}
& 15.1 & 50.8 & 58.3 & 41.4 & 43.5
& 19.5 & 50.2 & 60.5 & 42.5 & 45.5
& 44.2 & +2.1 & +2.8 \\
\rowcolor{gray!20}
\textbf{Ours} & \textbf{Proposed}
& \textbf{25.9} & 51.6 & 53.2 & \textbf{40.3} & 43.9
& \textbf{14.9} & \textbf{50.1} & \textbf{51.4} & \textbf{35.6} & \textbf{39.4}
& \textbf{19.1} & \textbf{47.9} & \textbf{48.5} & \textbf{32.5} & \textbf{38.2}
& \textbf{40.5} &\textbf{-3.4}  &\textbf{+6.5} \\
\bottomrule
\end{tabular}
}
\caption{Classification error rates (\%) for the ImageNet-to-ImageNet\texttt{++} CTTA task.}
\label{tab:ImageNet++}
\end{table*}

\section{Experiment}
\subsection{Experimental Setting}
\paragraph{Dataset.}
We conduct experiments on both image classification and semantic segmentation tasks. For classification, we use CIFAR100-to-CIFAR100C~\cite{krizhevsky2009learning} and ImageNet-C~\cite{hendrycks2019benchmarking}.
We further evaluate on our proposed Continual Repeated Shifts (CRS) benchmark, with two extended versions: ImageNet+ and ImageNet++, constructed by cyclically combining four ImageNet-derived datasets: ImageNet-V2~\cite{recht2019imagenet}, ImageNet-A~\cite{hendrycks2021natural}, ImageNet-R~\cite{hendrycks2021many}, and ImageNet-S~\cite{wang2019learning}.
For segmentation, we evaluate our method on Cityscapes-to-ACDC~\cite{cordts2016cityscapes,sakaridis2021acdc}.

\paragraph{CTTA Task Setting.}
Following CoTTA~\cite{wang2022continual}, for CIFAR100C and ImageNet-C, we evaluate on the largest corruption severity (level 5) and process all 15 corruption types sequentially as distinct domains. For ACDC, to reflect realistic temporal changes, we perform continual adaptation by looping through ACDC’s subdomains (Fog → Night → Rain → Snow) in three cycles.

On the CRS benchmark, for ImageNet+, we define four target domains: ImageNet-V2, ImageNet-A, ImageNet-R and ImageNet-S, and adapt in the fixed order (V2 → A → R → S), repeated three times. For ImageNet++, we employ a finer-grained schedule to create 12 domain shifts over three rounds. In each round, we use different subsets of the four datasets. More details are provided in the Appendix~\ref{appendix:imagenet++_construction}. We further introduce a benchmark-specific metric, \emph{Repeat Forget (RF)}, to quantify the degree of forgetting on repeated domains. For error-based metrics, it is defined as \( \mathrm{RF} = \text{Error}_{\text{final}} - \text{Error}_{\text{first}} \), where \(\text{Error}_{\text{first}}\) and \(\text{Error}_{\text{final}}\) are the average errors during the first and final rounds. A higher RF implies greater forgetting. For accuracy-based metrics, the sign is reversed.

\paragraph{Implementation Details.}
We adopt standardized CTTA settings for fair comparison. ViT-Base is used as the backbone for classification, and SegFormer-B5 for segmentation. The Adam optimizer and task-specific learning rates are used throughout. More hyperparameter and architecture details can be found in Appendix~\ref{appendix:impl}.

\begin{table*}[tb]                
\centering

\setlength{\tabcolsep}{1.5pt}       
{\small
\begin{tabular}{c|c|ccccc|ccccc|ccccc|c|c|c}
\toprule
\multicolumn{2}{c|}{Time} & \multicolumn{15}{c}{$t$ \makebox[10cm]{\rightarrowfill}} \\
\midrule
\multicolumn{2}{c|}{Round} &
\multicolumn{5}{c|}{1} & \multicolumn{5}{c|}{2} & \multicolumn{5}{c|}{3} &
\multicolumn{1}{c|}{\multirow{2}{*}{Mean$\uparrow$}} & 
\multicolumn{1}{c}{\multirow{2}{*}{RF$\downarrow$}} & 
\multicolumn{1}{|c}{\multirow{2}{*}{Gain}} \\
\cmidrule{1-17}
Method & REF & Fog & Night & Rain & Snow & Mean$\uparrow$ &
Fog & Night & Rain & Snow & Mean$\uparrow$ &
Fog & Night & Rain & Snow & Mean$\uparrow$ & & \\ 
\midrule
Source & ICLR2021
& 69.1 & 40.3 & 59.7 & 57.8 & 56.7
& 69.1 & 40.3 & 59.7 & 57.8 & 56.7
& 69.1 & 40.3 & 59.7 & 57.8 & 56.7
& 56.7 & 0.0 & / \\

TENT & ICLR2021
& 69.0 & 40.2 & 60.1 & 57.3 & 56.7
& 68.3 & 39.0 & 60.1 & 56.3 & 55.9
& 67.5 & 37.8 & 59.6 & 55.0 & 55.0
& 55.7 & +1.7 & -1.0 \\

CoTTA & CVPR2022
& 70.9 & 41.2 & 62.4 & 59.7 & 58.6
& 70.9 & 41.1 & 62.6 & 59.7 & 58.6
& 70.9 & 41.0 & 62.7 & 59.7 & 58.6
& 58.6 & 0.0  & +1.9 \\

SVDP & AAAI2024
& 72.1 & 44.0 & 65.2 & 63.0 & 61.1
& 72.2 & 44.5 & 65.9 & 63.5 & 61.5
& 72.1 & 44.2 & 65.6 & 63.6 & 61.4
& 61.3 & -0.3 & +4.6 \\ 

BECoTTA & ICML2024
& 71.5 & 42.6 & 63.2 & 59.1 & 59.1
& 71.5 & 42.6 & 63.2 & 59.1 & 59.1
& 71.5 & 42.5 & 63.2 & 59.1 & 59.1
& 59.1 & 0.0  &+2.4 \\ 
ADMA &CVPR2024
&71.9 &44.6 &67.4 &63.2 &61.8 
&71.7 &44.9 &66.5 &63.1 &61.6 
&72.3 &45.4 &67.1 &63.1 &62.0 
&61.8 & -0.2  &+5.1 \\

\rowcolor{gray!20}
\textbf{Ours} & \textbf{Proposed}
& \textbf{72.6} & \textbf{44.2} & \textbf{67.0} & \textbf{64.4} & \textbf{62.1}
& \textbf{73.2} & \textbf{45.5} & \textbf{68.0} & \textbf{64.9} & \textbf{62.9}
& \textbf{73.2} & \textbf{45.6} & \textbf{68.2} & \textbf{65.2} & \textbf{63.1}
& \textbf{62.7} &\textbf{-1.0}  & \textbf{+6.0} \\
\bottomrule
\end{tabular}
}
\caption{Performance comparison for Cityscapes-to-ACDC CTTA. “Mean” is the average mIoU.}
\label{tab:acdc}
\end{table*}

\subsection{Classification CTTA Tasks}

We first evaluate our proposed framework on the challenging ImageNet-C and our CRS benchmark. For completeness, CIFAR100-C results are provided in Appendix~\ref{cifar-c}.

\paragraph{Results on ImageNet-C.}
As shown in Table~\ref{tab:ImageNet-C}, our method lowers the error to 40.2\%, delivering a 15.6\% gain over the frozen source model and outperforming the previous best result (ADMA, 42.5\%) by 2.3\%. It achieves the lowest error on 10 of the 15 corruption types, including the challenging \textit{Defocus}, \textit{Glass~Blur}, \textit{Fog}, and \textit{JPEG}, demonstrating that our
dual-branch expert architecture, combined with the FDD mechanism, effectively removes corruption-specific noise while preserving task-relevant semantics.

\paragraph{Results on ImageNet+.}
Table~\ref{tab:ImageNet+} presents results under our CRS benchmark, where four domains (V2, A, R, S) recur over three rounds. Our method achieves a mean error of 44.5\%, outperforming the source model by 6.7\% and CoTTA by 5.7\%. While ViDA performs well initially (error 47.0\% vs.\ 46.5\% for ours in round 1), it suffers from severe forgetting in subsequent rounds, yielding a high Repeat Forget (RF) of +5.0. In contrast, our method not only maintains stable performance (44.0\%, 43.1\%) but also achieves an RF of -3.4, demonstrating strong resistance to forgetting. 

\paragraph{Results on ImageNet++.}
The finer-grained ImageNet++ benchmark (Table~\ref{tab:ImageNet++}) presents a more challenging CRS setting with 12 structured shifts. Our method achieves the lowest overall error of 40.5\%, outperforming the ViDA by 3.7\%. Unlike ViDA, which shows a clear performance drop in later rounds (RF = +2.1), our method consistently improves over time, achieving an RF of -3.4. This highlights our model’s ability to adapt to new domains while retaining and refining knowledge under repeated shifts.

\subsection{Semantic Segmentation CTTA Task}
On the Cityscapes-to-ACDC CTTA benchmark, as shown in Table~\ref{tab:acdc}, our method achieves the best performance with an average mIoU of 62.7\%, outperforming all baselines. It maintains consistent gains across all rounds without performance degradation. Compared to recent methods such as ADMA and SVDP, our approach demonstrates superior robustness and adaptability, highlighting the effectiveness of our dual-expert framework and frequency-aware adaptation.

\subsection{Ablation Study}
\paragraph{Expert Module Ablation.}
We ablate the dual‑branch expert design by selectively enabling two branches (see Table~\ref{tab:imagenet_ablation}). Without any experts, the backbone struggles on all three benchmarks. Enabling only the shared branch reduces corruption error on ImageNet-C but brings limited gains on the ImageNet+ and ImageNet++. In contrast, Activating only the domain self-adaptive branch cuts error more uniformly, confirming the benefit of per‑domain specialization under large shifts. 
Combining both branches yields the lowest error across all benchmarks, as the shared expert provides stable, reusable semantics, while the self-adaptive experts refine domain-specific residuals. The two are thus complementary, not redundant.

\begin{table}[t]
\centering
\setlength{\tabcolsep}{1.1mm}
\small{
\begin{tabular}{cc|c@{\hskip 3pt}c@{\hskip 3pt}c}
\toprule
Shared & Self-Adaptive & ImageNet-C & ImageNet+ & ImageNet++ \\
\midrule
 &  & 55.8 & 51.2 & 47.0 \\
\checkmark &  & 43.1 & 49.7 & 45.8 \\
 & \checkmark & 41.3 & 46.3 & 42.7 \\
\checkmark & \checkmark & \textbf{40.2} & \textbf{44.5} & \textbf{40.5} \\
\bottomrule
\end{tabular}
}
\caption{Ablation of the expert branch. Numbers are classification error rates (lower is better).}
\label{tab:imagenet_ablation}
\end{table}

\begin{table}[t]
  \centering
  \renewcommand{\arraystretch}{1.0}
  {\small
  \begin{tabular}{@{}lccc@{}}
    \toprule
    Variant & $K$ & Error $\downarrow$ & Diff $\downarrow$ \\
    \midrule
    Upper (oracle) & 15 & 39.7 & -- \\
    \midrule  
    Conf-Jump (AAAI2024)      & 26 & 50.2 & 10.5 \\
    FDD w/o covariance      & 12 & 41.3 & 1.6 \\
    FDD w/o discriminator         & 7  & 46.6 & 3.9 \\
    FDD w/o robust online update             & 7  & 41.8 & 2.1 \\
    FDD (ours)    & 7  & 40.2 & 0.5 \\
    \bottomrule
  \end{tabular}
  }
  \caption{Ablation of the FDD; $K$ denotes the number of discovered domains, i.e., the corresponding domain self-adaptive experts created; Diff is the error gap to Upper.}
  \label{tab:sodd-ablation}
\end{table}

\begin{figure*}[tb]   
  \centering
  \includegraphics[width=0.3\textwidth]{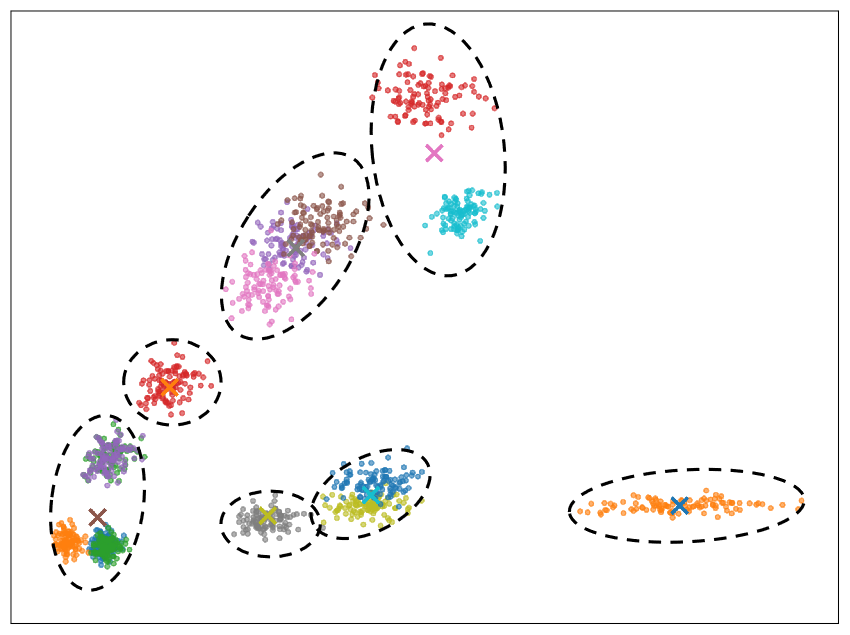}\hfill
  \includegraphics[width=0.3\textwidth]{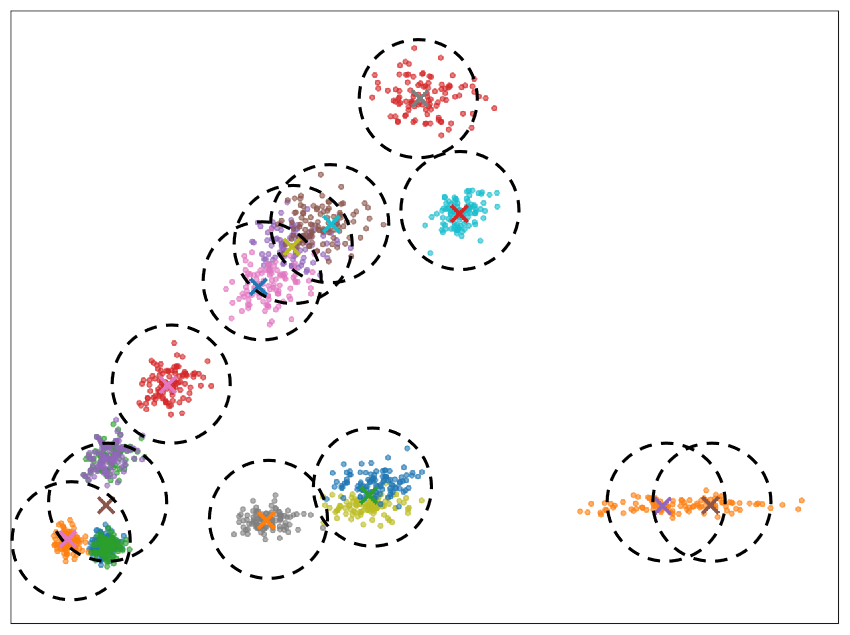}\hfill
  \includegraphics[width=0.3\textwidth]{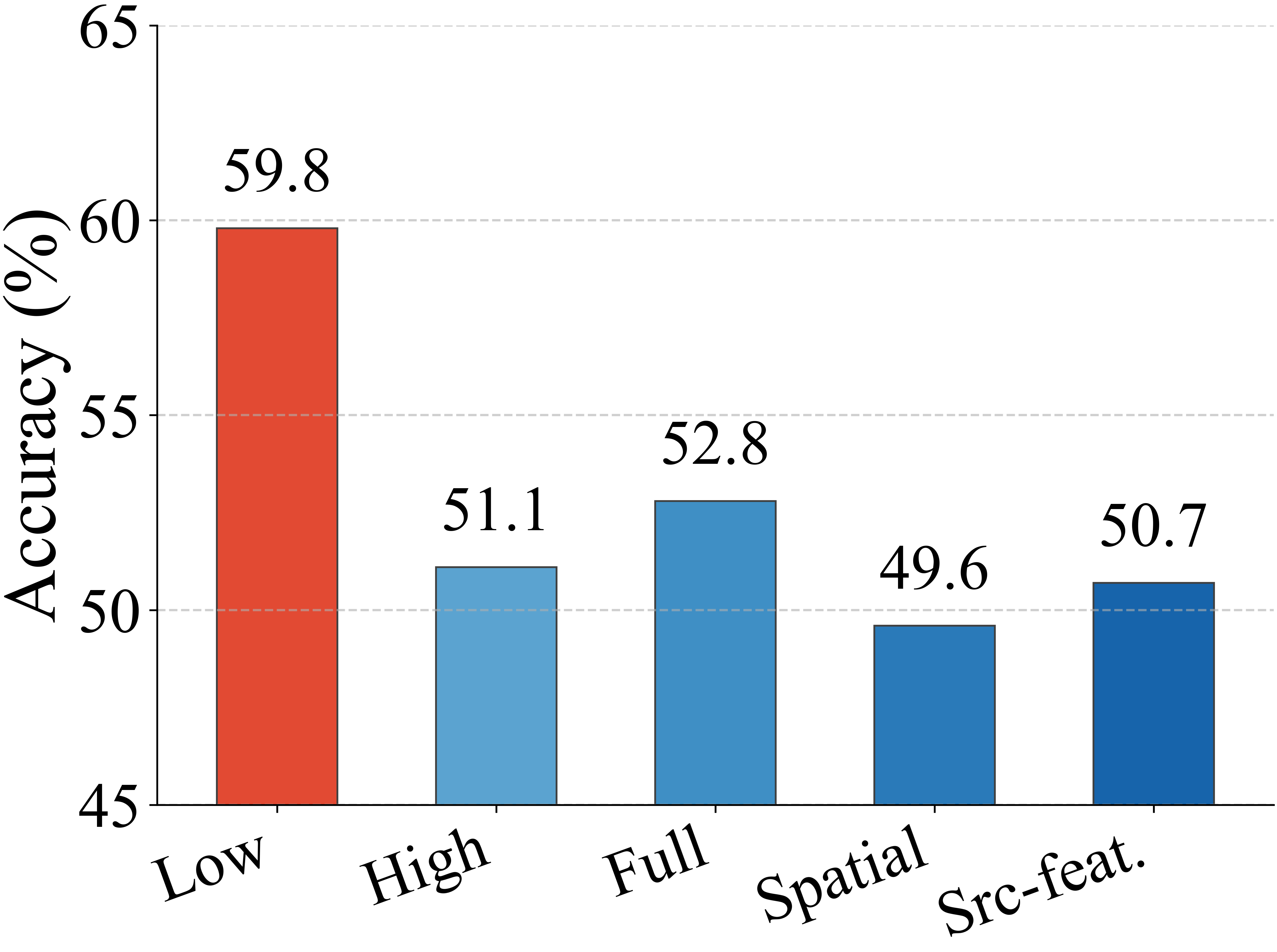}

  \makebox[0.3\textwidth]{\small (a) FDD(ours)}\hfill
  \makebox[0.3\textwidth]{\small (b) FDD w/o covariance}\hfill
  \makebox[0.3\textwidth]{\small (c) Feature-Selection Ablation}

  \caption{\textbf{FDD visualization and feature ablation on ImageNet‑C.}
           (a) PCA projection of low‑frequency features with full FDD; six well‑separated clusters are observed, “$\times$” denote estimated domain centres $\mu_i$.
           (b) Same projection with FDD without covariance, showing overlapping clusters and reduced separability.
           (c) Feature-Selection Ablation showing classification accuracy on ImageNet-C when the default low-frequency descriptor is replaced with four alternatives.}
  \label{fig:sodd-three-up}
\end{figure*}

\paragraph{Ablation Study on FDD.}
We conduct ablations on ImageNet-C to evaluate the effectiveness of FDD, as summarized in Table~\ref{tab:sodd-ablation}.
The \textit{Upper (oracle)} variant assumes known domain boundaries and achieves the lowest error.
Our full \textit{FDD} detects 7 domains (Fig.~\ref{fig:sodd-three-up}(a)) and achieves nearly optimal performance with far fewer experts, showing its efficiency.
\textit{Conf-Jump} adds a new expert based on confidence spikes~\cite{gan2023decorate}, leading to 26 experts and 50.2\% error, showing its unreliability for domain detection.
\textit{FDD w/o covariance} uses Euclidean distance instead of Mahalanobis, which blurs domain boundaries (Fig.~\ref{fig:sodd-three-up}(b)) and increases error to 41.3\%.
\textit{FDD w/o discriminator} disables expert selection and routes inputs randomly, resulting in the worst performance among all variants.
\textit{FDD w/o Robust Online Update} replaces our weighted update with EMA, increasing error to 41.8\%.
These results highlight the necessity of each component in FDD to achieve stable and accurate domain shift detection.

\paragraph{Feature-Selection Ablation.}
Fig.~\ref{fig:sodd-three-up}(c) reports accuracy after replacing our default low-frequency descriptor with four alternatives. Using high-frequency (High), full-spectrum (Full), spatial RGB histogram (Spatial), or ViT-CLS token representations (Src-feat.) all leads to performance drops. These results reinforce that the raw low-frequency patch achieves the best balance, capturing shift-specific global statistics while suppressing category details for precise and robust discrimination.

\section{Conclusion}
In this paper, we propose a frequency-discriminative shared \& self-adaptive domain expert model that achieves a strong balance between adaptation and forgetting in CTTA. Under the proposed CRS benchmark with repeated and diverse domain shifts, our method demonstrates consistent performance improvement while avoiding catastrophic forgetting. These results highlight its practicality in realistic, non-stationary environments.
\paragraph{Limitation.}
Our current design adopts a fixed fusion strategy between shared and domain self-adaptive experts. While effective, this static combination may limit adaptability under more complex or dynamic shifts. Future work could explore learnable or input-aware fusion mechanisms to further enhance flexibility and robustness.


\section*{Acknowledgments}
 This work is supported by the National Natural Science Foundation of China under Grant No.U21B2048 and No.62302382, the Shenzhen Key Technical Projects Under Grant CJGJZD20220517141605013, and the China Postdoctoral Science Foundation No.2024M752584.

\bibliography{aaai2026}

\newpage
\clearpage
\appendix
\setcounter{figure}{0}
\setcounter{table}{0}
\setcounter{equation}{0}
\appendix

\section{Visualization of Expert Attention Across Target Domains}
\label{app:vis}

\begin{figure*}[t]
    \centering
    \includegraphics[width=\textwidth]{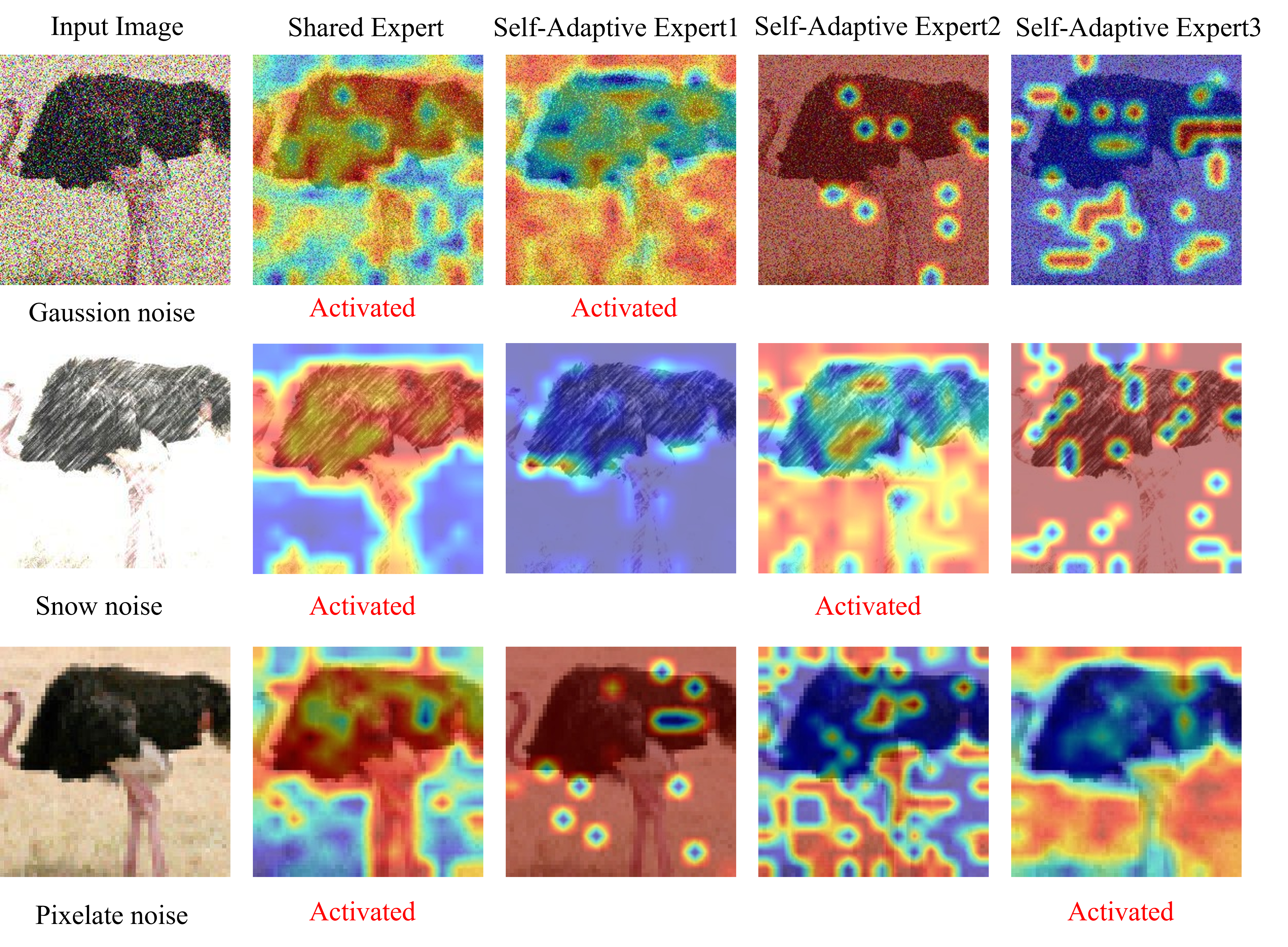}
    \caption{Visualization of the \textbf{domain-shared expert} and the three
    \textbf{domain self-adaptive experts} on three corruption domains
    (\emph{Gaussian}, \emph{Snow}, and \emph{Pixelate}).  
    Each panel is an \emph{Attention Map}, obtained by measuring the
    similarity between the \texttt{[CLS]} token and all patch tokens produced by the corresponding expert branch; brighter colours denote stronger
    affinity. The column labelled “Activated’’ indicates that the expert was selected
    by the Frequency-aware Domain Discriminator (FDD) for the current input, whereas the remaining experts were inactive.}
    \label{fig:vis_appendix}
\end{figure*}

Figure~\ref{fig:vis_appendix} illustrates how different expert branches allocate attention when confronted with three representative corruption domains.  It can be observed that  
(1) the shared expert consistently focuses on stable semantic regions
such as the ostrich’s head and torso, demonstrating domain-invariant
behaviour;  
(2) the single domain self-adaptive expert activated by the FDD shifts
its attention towards corruption-specific artefacts—grainy blobs for
\emph{Gaussian}, elongated streaks for \emph{Snow}, and block boundaries
for \emph{Pixelate}—capturing complementary domain-specific cues;  
and (3) the two non-activated experts produce diffuse or fragmented maps with
no clear semantic correspondence, indicating a mismatch between their
feature spaces and the current input domain.  
These visual findings prove that the shared expert provides a stable
semantic scaffold, while domain-adaptive experts inject domain-aware details on demand, achieving efficient and robust adaptation.
\section{Derivation of the Bayesian Posterior with Shrinkage-Regularized Mahalanobis Distance}
\label{app:mahalanobis}
Given an input embedding $z \in \mathbb{R}^{d_f}$, and assuming for each domain $i$ that the low-frequency descriptors follow a multivariate Gaussian distribution:
\begin{equation}
p(z \mid y=i) \;=\; \mathcal{N}\!\bigl(z \,\bigl|\, \mu_i,\, \Sigma_i\bigr),
\label{eq:gaussian}
\end{equation}
where $\mu_i$ and $\Sigma_i$ are the mean and covariance of domain $i$.
The explicit form of the multivariate Gaussian density is:
\begin{equation}
p(z \mid y=i)
  = \frac{1}{c_i}\,
    \exp\!\Bigl(
      -\tfrac12 (z-\mu_i)^{\!\top}\Sigma_i^{-1}(z-\mu_i)
    \Bigr),
\label{eq:gaussian_density}
\end{equation}
where \(c_i = (2\pi)^{d_f/2}\lvert\Sigma_i\rvert^{1/2}\).
We also assume a uniform prior over domains: $P(y=i) = 1/K$.

According to Bayes’ rule, the posterior probability of domain $i$ given $z$ is:
\begin{equation}
P(y=i \mid z)
= \frac{p(z \mid y=i) \, P(y=i)}
       {\displaystyle\sum_{j=1}^{K} p(z \mid y=j)\, P(y=j)}.
\label{eq:bayes_general}
\end{equation}
Because the prior is uniform, this simplifies to:
\begin{equation}
P(y=i \mid z)
= \frac{p(z \mid y=i)}
       {\displaystyle\sum_{j=1}^{K} p(z \mid y=j)}.
\label{eq:bayes_uniform}
\end{equation}

Plugging \eqref{eq:gaussian_density} into \eqref{eq:bayes_uniform} and
cancelling the common constant $(2\pi)^{-d_f/2}$ yields
{\small{
\begin{equation}
P(y=i \mid z)
= \frac{
        \exp\!\bigl(-\tfrac12 (z - \mu_i)^{\!\top}\Sigma_i^{-1}(z - \mu_i)\bigr) /
        \lvert \Sigma_i \rvert^{1/2}
      }{
        \displaystyle\sum_{j=1}^{K}
        \exp\!\bigl(-\tfrac12 (z - \mu_j)^{\!\top}\Sigma_j^{-1}(z - \mu_j)\bigr) /
        \lvert \Sigma_j \rvert^{1/2}
      }.
\label{eq:posterior_mahal_basic}
\end{equation}
}}
To improve numerical stability, we introduce a shrinkage-regularised
Mahalanobis distance
\begin{equation}
m_i(z)
\;=\;
\bigl(z - \mu_i\bigr)^{\!\top}
\Bigl[(1 - \varepsilon)\,\Sigma_i + \varepsilon I\Bigr]^{-1}
\bigl(z - \mu_i\bigr),
\label{eq:shrinkage_mahal}
\end{equation}
where \(I\) is the identity matrix and \(\varepsilon\in[0,1)\) is a
shrinkage parameter that blends the sample covariance with the identity
to mitigate ill-conditioning.

Substituting $m_i(z)$ into \eqref{eq:posterior_mahal_basic} gives the
final form of the posterior:
\begin{equation}
P(y=i \mid z)
= \frac{
        \exp\!\bigl[-\tfrac12 m_i(z)\bigr] /
        \sqrt{\det \Sigma_i}
      }{
        \displaystyle\sum_{j=1}^{K}
        \exp\!\bigl[-\tfrac12 m_j(z)\bigr] /
        \sqrt{\det \Sigma_j}
      }.
\label{eq:posterior_final}
\end{equation}

This completes the derivation.





\section{Equivalence of Posterior Maximization and Mahalanobis Distance Minimization}
\label{app:posterior-mahalanobis}

The posterior probability of domain $i$ given a sample embedding $z$ is defined as:
\begin{equation}
P(y=i \mid z)
= \frac{%
      \exp\!\bigl[ -\tfrac12\, m_i(z) \bigr] \, / \sqrt{\det\Sigma_i}%
    }{%
      \displaystyle\sum_{j=1}^{K}
      \exp\!\bigl[ -\tfrac12\, m_j(z) \bigr] \, / \sqrt{\det\Sigma_j}%
    },
\label{eq:posterior}
\end{equation}
where $m_i(z)$ is the regularized Mahalanobis distance:
\begin{equation}
m_i(z)=
(z-\mu_i)^{\!\top}\!
\bigl[(1-\varepsilon)\Sigma_i+\varepsilon I\bigr]^{-1}
(z-\mu_i).
\label{eq:mahalanobis}
\end{equation}
and ignoring the denominator (which is constant across $i$ in $\arg\max$), we get:
\begin{equation}
\log P(y=i \mid z)
\propto
-\tfrac12\, m_i(z)\;-\;\tfrac12\log\det\Sigma_i.
\label{eq:log-posterior}
\end{equation}
Thus, the posterior is maximized when the following is minimized:
\begin{equation}
m_i(z)\;+\;\log\det\Sigma_i
\label{eq:min-criterion}
\end{equation}

Considering that the Mahalanobis distance is the dominant factor in practice, we omit the influence of $\det \Sigma_i$ and approximate the MAP decision rule as:
\begin{equation}
\arg\max_{i} P(y=i \mid z)\;\;\approx\;\;\arg\min_{i} m_i(z),
\label{eq:map-rule}
\end{equation}

\section{Derivation of Online Updates for $\mu_{i^*}^{\text{new}}$ and $\Sigma_{i^*}^{\text{new}}$}
\label{app:mle-update}

\subparagraph{Weighted complete-data log-likelihood.}
Let $\mathcal{D}_{\text{past}}^{(i^*)}$ denote all batch samples that have
already been assigned to domain~$i^*$ and let its current cumulative weight (or effective
 batchcount) be $c_{i^*}=|\mathcal{D}_{\text{past}}^{(i^*)}|$.
For the current mini-batch
$\mathcal{B}=\{x_b\}_{b=1}^{B}$ we extract low-frequency embeddings
$z_b=f(x_b)\in\mathbb{R}^{d_f}$.
The domain discriminator supplies soft responsibilities
$w_b=P(y=i^*\mid z_b)$ with $\sum_{b=1}^{B}w_b=1$.
We further keep the Gaussian statistics
$(\mu_{i^*},\Sigma_{i^*})$ estimated from
$\mathcal{D}_{\text{past}}^{(i^*)}$.

\textbf{Step 1 — Joint likelihood.}\;
Assuming every feature in domain~$i^*$ is i.i.d.\ as
$z\sim\mathcal{N}(\mu,\Sigma)$, 
the joint likelihood of the \emph{augmented} data set
$\mathcal{D}_{\text{past}}^{(i^*)}\cup\mathcal{B}$ is proportional to
\[
\prod_{k=1}^{c_{i^*}}\!\mathcal{N}\!\bigl(z^{(k)}\mid\mu,\Sigma\bigr)
\;\prod_{b=1}^{B}\!\mathcal{N}\!\bigl(z_b\mid\mu,\Sigma\bigr)^{w_b},
\]
where $\{z^{(k)}\}$ are the $c_{i^*}$ historical features.

\textbf{Step 2 — Virtual-sample approximation.}\;
Replacing each historical feature $z^{(k)}$ by the empirical mean
$\mu_{i^*}$—a standard sufficient-statistic shortcut—converts the first
product into the $c_{i^*}$-fold power of a \emph{single} likelihood term:
\[
\bigl[\mathcal{N}\!\bigl(\mu_{i^*}\mid\mu,\Sigma\bigr)\bigr]^{c_{i^*}}
\;\prod_{b=1}^{B}\!\mathcal{N}\!\bigl(z_b\mid\mu,\Sigma\bigr)^{w_b}.
\]

\textbf{Step 3 — Take logarithms.}\;
Using $\log(ab)=\log a+\log b$ and $\log a^{c}=c\log a$ we obtain the
\textbf{weighted complete-data log-likelihood}
\begin{equation}
\mathcal{L}(\mu,\Sigma)=
c_{i^*}\,\log\mathcal{N}\!\bigl(\mu_{i^*}\mid\mu,\Sigma\bigr)
+\sum_{b=1}^{B}w_b\,\log\mathcal{N}\!\bigl(z_b\mid\mu,\Sigma\bigr).
\label{eq:weighted-llh}
\end{equation}
In the main text we write the \textbf{first term}
$c_{i^*}\log\mathcal{N}(\mu_{i^*}\mid\mu,\Sigma)$ more compactly as
$c_{i^*}\log\mathcal{N}(\mu_{i^*},\Sigma_{i^*})$
to emphasise its origin in the historical statistics
$(\mu_{i^*},\Sigma_{i^*})$.

\subparagraph{Gaussian log-density.}  
For any $z\in\mathbb{R}^{d_f}$,
\begin{equation}
\begin{aligned}
\log \mathcal{N}(z \mid \mu, \Sigma)
&= -\frac{d_f}{2}\log(2\pi) -\frac{1}{2}\log \det \Sigma \\
&\quad -\frac{1}{2}(z-\mu)^{\!\top} \Sigma^{-1} (z-\mu)
\end{aligned}
\label{eq:gauss-log}
\end{equation}

\subsection{Update for the mean $\boldsymbol{\mu_{i^*}^{\text{new}}}$}
Substituting \eqref{eq:gauss-log} into \eqref{eq:weighted-llh} and
collecting terms that depend on~$\mu$ give
\begin{equation}
\begin{aligned}
\mathcal{L}_{\mu}
&= 
-\frac{1}{2} c_{i^*} (\mu_{i^*} - \mu)^{\!\top} \Sigma^{-1} (\mu_{i^*} - \mu) \\
&\quad
-\frac{1}{2} \sum_{b=1}^{B} w_b (z_b - \mu)^{\!\top} \Sigma^{-1} (z_b - \mu)
\end{aligned}
\label{eq:mean-part}
\end{equation}
Setting $\partial\mathcal{L}_{\mu}/\partial\mu=0$ and premultiplying by
$\Sigma$ yield
\[
c_{i^*}(\mu_{i^*}-\mu)+\sum_{b=1}^{B}w_b(z_b-\mu)=0.
\]
Because $\sum_{b}w_b=1$, solving for $\mu$ gives
\begin{equation}
\boxed{
\mu_{i^*}^{\text{new}}=
\frac{c_{i^*}\,\mu_{i^*}+\displaystyle\sum_{b=1}^{B}w_b z_b}{c_{i^*}+1}.
}
\label{eq:mu-update}
\end{equation}
Hence the new mean is a convex combination of the
historical mean and the weighted batch mean.

\subsection{Update for the covariance $\boldsymbol{\Sigma_{i^*}^{\text{new}}}$}
Keeping only the $\Sigma$-dependent terms in \eqref{eq:weighted-llh}
gives
\begin{align}
\mathcal{L}_{\Sigma}
&= -\frac12(c_{i^*}+1)\log\!\det\Sigma \notag\\
&\quad -\frac12\,c_{i^*}(\mu_{i^*}-\mu_{i^*}^{\text{new}})^{\!\top}
          \Sigma^{-1}(\mu_{i^*}-\mu_{i^*}^{\text{new}}) \notag\\
&\quad -\frac12\sum_{b=1}^{B}w_b
          (z_b-\mu_{i^*}^{\text{new}})^{\!\top}\Sigma^{-1}
          (z_b-\mu_{i^*}^{\text{new}}).
\label{eq:cov-part}
\end{align}

Taking $\partial\mathcal{L}_{\Sigma}/\partial\Sigma^{-1}=0$ yields the
scatter-matrix condition, from which we obtain the \textbf{exact update}
\begin{equation}
\label{eq:sigma-exact}
\begin{aligned}
\Sigma_{i^*}^{\text{exact}}
&= \frac{1}{c_{i^*}+1}
\Bigl(
      c_{i^*}\Bigl[\Sigma_{i^*}
        +(\mu_{i^*}-\mu_{i^*}^{\text{new}})
         (\mu_{i^*}-\mu_{i^*}^{\text{new}})^{\!\top}\Bigr] \\
&\hspace{4.5em}
      +\sum_{b=1}^{B}
        w_b\,(z_b-\mu_{i^*}^{\text{new}})
             (z_b-\mu_{i^*}^{\text{new}})^{\!\top}
\Bigr).
\end{aligned}
\end{equation}

\paragraph{Approximation.}
In practice the mean shift
$(\mu_{i^*}-\mu_{i^*}^{\text{new}})$ is typically small when
$c_{i^*}$ is large (most past data), or when the new batch carries
limited weight.  Therefore, we approximate $\mu_{i^*}^{\text{new}}$ by $\mu_{i^*}$ to reduce computational overhead, which yields the simplified update adopted in the main text:
\begin{equation}
\boxed{
\Sigma_{i^*}^{\text{new}}=
\frac{
      c_{i^*}\Sigma_{i^*}
      +\displaystyle\sum_{b=1}^{B}
        w_b\,(z_b-\mu_{i^*})
            (z_b-\mu_{i^*})^{\!\top}}
     {c_{i^*}+1}.
}
\label{eq:sigma-approx}
\end{equation}

\section{Construction Details of ImageNet+ and ImageNet++}
\label{appendix:imagenet++_construction}
\subsection{ImageNet+ Construction}

ImageNet+ is designed to simulate realistic, large-scale, and recurring domain shifts within the CRS benchmark. Specifically, we define four target domains: ImageNet-V2, ImageNet-A, ImageNet-R, and ImageNet-S. During evaluation, adaptation proceeds in a fixed order (V2 $\rightarrow$ A $\rightarrow$ R $\rightarrow$ S), which constitutes one complete cycle. This sequence is repeated three times, resulting in a total of 12 domain shifts. In each cycle, the full set of images from each dataset is used, i.e., every image in each domain is included in every round. This protocol ensures consistent and comprehensive exposure to each domain, faithfully mimicking recurring shifts in real-world environments.

\subsection{ImageNet++ Construction}

To further increase the granularity and difficulty of the continual adaptation task, ImageNet++ adopts a more fine-grained protocol for the CRS benchmark. Similar to ImageNet+, the evaluation consists of three rounds, each traversing the four domains in the fixed order (V2 $\rightarrow$ A $\rightarrow$ R $\rightarrow$ S), again resulting in 12 domain shifts in total. 

However, in ImageNet++, different and non-overlapping subsets of the four datasets are used in each round to simulate evolving but recurring domains:
\begin{itemize}
    \item \textbf{ImageNet-V2:} For each round, a unique subset of ImageNet-V2 is sampled. These subsets are constructed using different strategies (such as matched-frequency, threshold-0.7, or top-images), and are strictly non-overlapping across the three rounds. This ensures that all images from ImageNet-V2 are seen exactly once during the entire evaluation process.
    \item \textbf{ImageNet-A:} Due to its relatively small scale, the complete ImageNet-A dataset is used in every round.
    \item \textbf{ImageNet-R and ImageNet-S:} For both domains, non-overlapping subsets are constructed for each round, ensuring complete coverage of all images across the three rounds.
\end{itemize}

Overall, this protocol guarantees that every image in ImageNet-V2, ImageNet-R, and ImageNet-S is evaluated exactly once, while ImageNet-A is reused in each cycle. By employing non-overlapping subsets for each round, ImageNet++ creates a more challenging and realistic continual adaptation benchmark, as it better reflects recurring yet dynamically changing domain distributions often observed in real-world applications.

\begin{table*}[ht]
\centering
\setlength\tabcolsep{2pt}
\resizebox{\linewidth}{!}{
\begin{tabular}{c|c|ccccccccccccccc|cc}
\toprule
Method & REF &
\rotatebox[origin=c]{50}{Gaussian} & \rotatebox[origin=c]{50}{Shot} & \rotatebox[origin=c]{50}{Impulse} & \rotatebox[origin=c]{50}{Defocus} & \rotatebox[origin=c]{50}{Glass} & \rotatebox[origin=c]{50}{Motion} & \rotatebox[origin=c]{50}{Zoom} & \rotatebox[origin=c]{50}{Snow} & \rotatebox[origin=c]{50}{Frost} & \rotatebox[origin=c]{50}{Fog} & \rotatebox[origin=c]{50}{Brightness} & \rotatebox[origin=c]{50}{Contrast} & \rotatebox[origin=c]{50}{Elastic} & \rotatebox[origin=c]{50}{Pixelate} & \rotatebox[origin=c]{50}{JPEG} & Mean$\downarrow$ & Gain \\
\midrule
Source~\cite{dosovitskiy2020image} & ICLR2021  &55.0&51.5&26.9&24.0&60.5&29.0&21.4&21.1&25.0&35.2&11.8&34.8&43.2&56.0&35.9&35.4&0.0\\
Pseudo-label~\cite{lee2013pseudo} & ICML2013 &53.8&48.9&25.4&23.0&58.7&27.3&19.6&20.6&23.4&31.3&11.8&28.4&39.6&52.3&33.9&33.2&+2.2\\
TENT-continual~\cite{wang2020tent} &ICLR2021&53.0&47.0&24.6&22.3&58.5&26.5&19.0&21.0&23.0&30.1&11.8&25.2&39.0&47.1&33.3&32.1&+3.3\\
CoTTA~\cite{wang2022continual} & CVPR2022 &55.0&51.3&25.8&24.1&59.2&28.9&21.4&21.0&24.7&34.9&11.7&31.7&40.4&55.7&35.6&34.8&+0.6\\
VDP~\cite{gan2023decorate} & AAAI2023 &54.8&51.2&25.6&24.2&59.1&28.8&21.2&20.5&23.3&33.8&\textbf{7.5}&\textbf{11.7}&32.0&51.7&35.2&32.0&+3.4\\

ViDA~\cite{vida} & ICLR2024 &50.1 & 40.7 & 22.0 & 21.2 & 45.2 & 21.6 & 16.5 & \textbf{17.9} & 16.6 & 25.6 & 11.5 & 29.0 & 29.6 & 34.7 & 27.1 & 27.3 & +8.1\\

ADMA~\cite{liu2024continual} & CVPR2024 & 48.6 &30.7 & \textbf{18.5} & 21.3 & 38.4 & 22.2 & 17.5 & 19.3 & 18.0 & 24.8 & 13.1 & 27.8 & 31.4 & 35.5 & 29.5& 26.4 &+9.0 \\

\rowcolor{gray!20}
\textbf{Ours} & \textbf{Proposed} & \textbf{43.1} &\textbf{30.3} & 19.0 & \textbf{19.6} & \textbf{37.2} & \textbf{21.4} & \textbf{16.2} & 18.4 & \textbf{15.5} & \textbf{22.7} & 11.9 & 19.6 & \textbf{27.3} & \textbf{25.8} & \textbf{28.9} & \textbf{23.8} & \textbf{+11.6} \\
\bottomrule
\end{tabular}
}
\caption{Classification error rates (\%) for the Cifar100-to-Cifar100C CTTA task. Mean indicates the average error across 15 corruption types. Gain represents the accuracy improvement over the source model.}
\label{tab:cifar100c}
\end{table*}

\section{Appendix: Implementation Details}
\label{appendix:impl}

\paragraph{Classification.}
We adopt ViT-Base~\cite{dosovitskiy2020image} as the backbone.
Images from CIFAR-100C are resized to $384\times384$, whereas those from
ImageNet-C, ImageNet+, and ImageNet++ are resized to
$224\times224$.

\paragraph{Segmentation.}
For semantic segmentation we use SegFormer-B5~\cite{xie2021segformer},
initialised with Cityscapes weights. ACDC images are down-sampled from
$1920\times1080$ to $960\times540$.

\paragraph{Optimisation.}
All networks are trained with Adam~\cite{kingma2014adam}
($\beta_1{=}0.9$, $\beta_2{=}0.999$).
The learning rate is set to
$1\!\times\!10^{-5}$ for CIFAR-100C,
$1\!\times\!10^{-5}$ for ImageNet-C,
$5\!\times\!10^{-4}$ for ImageNet+ and ImageNet++,
and $3\!\times\!10^{-4}$ for ACDC.
Unless stated otherwise, we use a batch size of 50, cosine-annealing
scheduling without warm-up, and a weight decay of $0.05$.

\paragraph{Model Hyperparameter and other Settings}
The Dual-Branch Expert Architecture is inserted in the feed-forward network (FFN) of each ViT block, with gating coefficient $\lambda = 0.5$.  
Low-rank factorisation uses rank $r{=}32$ for the shared branch and $r{=}16$ for each expert in a Mixture-of-Experts layer containing $M{=}2$ experts.  
The Frequency-Aware Domain Discriminator applies a frequency radiu $\ell = 16$; a decision threshold $\tau = 1.5$ is used to determine whether the current input belongs to a newly emerging domain, and a confidence threshold $\kappa = 0.4$ discards low-certainty predictions.  
To cut computation, we approximate each covariance matrix by its diagonal, $\Sigma_i \!\approx\! \operatorname{diag}\!\bigl(\sigma_{i,1}^2,\dots,\sigma_{i,d_f}^2\bigr)$, reducing GPU memory by roughly 60 \% on ImageNet++ without noticeable accuracy loss.

\paragraph{Runtime environment.}
Experiments are conducted on an Ubuntu 20.04 machine (kernel 5.15.0-70-generic),
equipped with four NVIDIA A800 80GB PCIe GPUs. All models are implemented in PyTorch~2.6.0\,+\,CUDA\,12.4.
During test, we use two GPUs and each GPU uses at most 15\,GB of memory,
and a full pass over a given dataset takes roughly one hour.
Every reported metric is the average of three independent runs
(seed~\{2025, 2026, 2027\}).

\section{Results on Cifar100-C}
\label{cifar-c}
As shown in Table~\ref{tab:cifar100c}, our method achieves the lowest average classification error rate of 23.8\% across all 15 corruption types on the Cifar100-to-Cifar100C CTTA benchmark, outperforming all previous state-of-the-art approaches. Compared to the strong baseline ADMA~\cite{liu2024continual} (26.4\%), our approach achieves a relative reduction of 2.6 percentage points in mean error, and brings a substantial gain of +11.6\% over the source model. Notably, our method obtains the best performance on the majority of corruption types, including Gaussian, Shot, Defocus, Glass, Motion, Zoom, Frost, Fog, Contrast, Elastic, Pixelate, and JPEG, demonstrating superior robustness and generalization under various distribution shifts. These improvements can be attributed to the proposed frequency-aware dual-branch expert framework and the online domain discriminator, which together enable effective suppression of domain-specific noise while preserving task-relevant features. Overall, the results highlight the effectiveness and practicality of our approach for continual test-time adaptation in dynamic environments.

\section{Parameter Efficiency of the Dual-Branch Expert Architecture.}
A key advantage of our dual-branch expert framework is its parameter efficiency. Specifically, the additional parameters introduced by our shared and domain self-adaptive experts constitute only 8\% of the total model parameters. Furthermore, thanks to the modular design and selective activation mechanism, at any given time during continual test-time adaptation, only the parameters of the currently selected domain self-adaptive expert are updated. As a result, the number of trainable parameters per adaptation step is as low as 1\% of the full model size, which is significantly lower than standard model-based CTTA approaches that require updating a large portion or even the entirety of the network. This lightweight design not only reduces memory and computational overhead but also mitigates the risk of overfitting and catastrophic forgetting. Empirical results confirm that our approach achieves strong adaptation performance with negligible parameter cost, making it highly suitable for resource-constrained and real-time deployment scenarios.

\end{document}